\documentclass[11pt]{article}

\usepackage[final]{acl}

\usepackage{times}
\usepackage{latexsym}

\usepackage[T1]{fontenc}

\usepackage[utf8]{inputenc}

\usepackage{microtype}

\usepackage{inconsolata}

\usepackage{graphicx}
\usepackage{amsmath}
\usepackage{multirow}
\usepackage{booktabs}
\usepackage{amssymb}
\usepackage{makecell}

\title{FinDialogLens: Event Extraction over Multi-Party Dialogue for Missed-Trade Identification in Financial Chatrooms}

\author{
Chin-Lun Fu~~~ Hong Ni~~~ Behrouz Madahian
  \\
  Machine Learning Center of Excellence, JPMorgan Chase \& Co.
  \\
  \small{
    \texttt{\{chin-lun.fu,hong.ni,behrouz.madahian\}@jpmchase.com}
  }
}

\begin{document}
\maketitle
\begin{abstract}
Multi-party financial chatrooms are vital for sales-and-trading professionals, but their complexity makes manual recovery of missed trades infeasible: each \emph{Request for Quote} (RFQ) is an event whose final price and trade outcome appear many messages after the \emph{RFQ-trigger message} (the inquiry message), interleaved with concurrent RFQs from other participants. We cast this as \textbf{event extraction (EE) over multi-party dialogue} and present \textbf{FinDialogLens}, a hybrid LLM pipeline in which compact fine-tuned classifiers act as inference-time scaffolds: they detect RFQ-triggers and price/trade outcome metadata, an RFQ-Level Module segments per-event RFQ windows, and a Trade Engine fills argument roles. With GPT-4o, FinDialogLens reaches 92.1\% and 94.3\% accuracy on final price and trade outcome, respectively, outperforming full-chatroom CoT prompting methods; fine-tuned open-source LLMs with as few as 3B parameters achieve comparable performance with modest in-domain data. To make the LLM-based solution practical at scale, a difficulty-aware router balances cost and accuracy by allocating RFQs between a low-cost rule-based engine and the higher-performing LLM-powered Trade Engine, cutting LLM calls by 85\% on final price while recovering half of the accuracy gap to FinDialogLens (GPT-4o), saving over \$300/day at our 70{,}000-RFQ/day scale.
\end{abstract}

\section{Introduction}
Financial institutions communicate with their clients at scale over multi-party chatrooms, where a single day routinely carries hundreds of messages and tens of distinct trade negotiations posted concurrently. A typical interaction begins when a client reaches out with a \emph{Request for Quote} (RFQ), e.g.,\ ``\textit{can I get a price quote on 5Y XYZ?}''; a sales representative responds with a quote (e.g.,``\textit{75 / 76}''), and the client either agrees (e.g.,``\textit{mine}'') or walks away (e.g.,``\textit{away, thanks}''). The inquiry, the final agreed price, and the trade outcome together form the structured record of the business event. In practice, RFQ negotiations are interleaved, asynchronous, and noisy: quotes, requotes, and trade outcomes may be separated by minutes, mixed with parallel client negotiations, and obscured by irrelevant market updates or chitchat. Consequently, identifying the trigger, quotes, and final outcome for a given RFQ requires linking distant, non-adjacent messages across substantial irrelevant context.

Executed trades are logged automatically by systems, but \emph{missed} trades, namely inquiries that did not result in execution, are not, leaving sales desks unable to measure quote competitiveness or prioritize follow-up. At production volumes of tens of thousands of RFQs per day, manual recovery of missed-trade records from chat is infeasible.

We cast this problem as \textbf{event extraction over multi-party dialogue}, an application of EE to a setting where each RFQ is an event whose \emph{trigger} is the inquiry message, and whose argument roles, the final price and the trade outcome, are filled by content from downstream messages that often appear many minutes later and are interleaved with unrelated concurrent RFQs from other participants. This combination of long-range argument scattering and multi-party interleaving is not the setting that existing EE recipes target: sentence-level methods~\citep{doddington-etal-2004-automatic, li-etal-2013-joint, chen-etal-2015-event} are bounded to a single sentence, while document-level methods~\citep{ebner-etal-2020-multi, li-etal-2021-document, zheng-etal-2019-doc2edag} assume coherent single-author text. Table~\ref{tab:ee-comparison} summarizes why other prior EE systems are not directly comparable.

\begin{figure*}[t]
  \centering
  \includegraphics[width=0.85\textwidth]{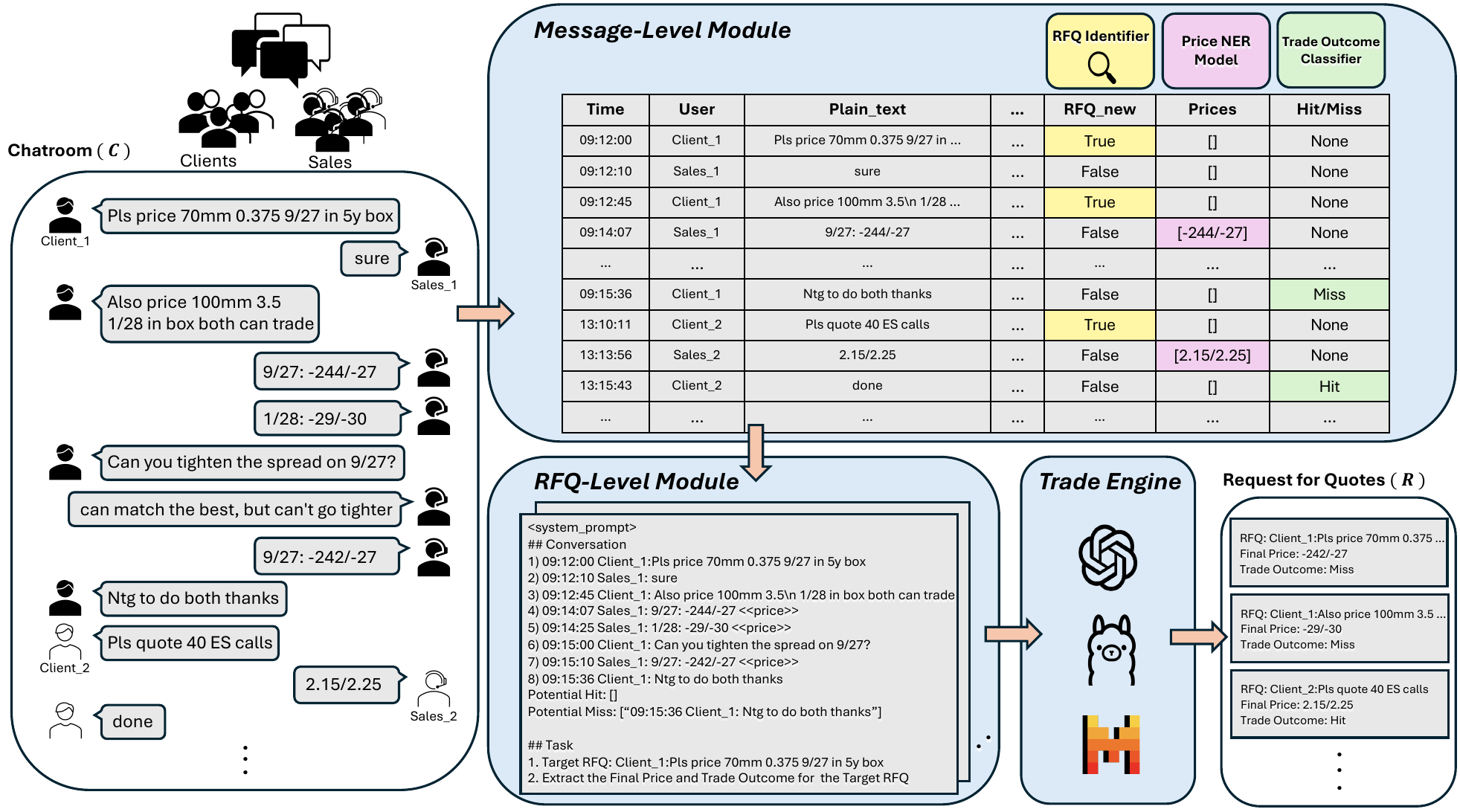}
\caption{Overview of the FinDialogLens pipeline for event extraction over multi-party dialogue. The Message-Level Module produces message-level metadata through RFQ-trigger detection, trade outcome classification, and price NER. The RFQ-Level Module uses these predictions to segment per-event RFQ windows from the chatroom. The Trade Engine performs LLM-based argument role filling over each RFQ window.}
  \label{fig:overview}
\end{figure*}

Our framework, \textbf{FinDialogLens} (Figure~\ref{fig:overview}), is a hybrid LLM pipeline with three modules. The \emph{Message-Level Module} runs compact RoBERTa-base~\citep{liu2019robertarobustlyoptimizedbert} classifiers fine-tuned on in-domain data to produce metadata for RFQ-trigger messages, price mentions, and trade outcome cues; these predictions act as \emph{inference-time scaffolds} for the downstream LLM, supplying domain-specific signals that general-purpose LLMs lack on raw chat~\citep{yu2023open}. The \emph{RFQ-Level Module} segments the chatroom into per-event RFQ windows, one for each detected RFQ-trigger message, using a temporal heuristic. The \emph{Trade Engine} performs LLM-based argument role filling over each RFQ window using GPT-4o~\citep{openai2024gpt4ocard} or a fine-tuned open-source LLM. To make the LLM-based solution practical at production scale, a \emph{difficulty-aware router} predicts per-event difficulty and assigns easier RFQ windows\footnote{Here, ``easier'' denotes RFQ windows that follow typical conversational patterns and can be addressed reliably using heuristic rules.} to a low-cost rule-based weak engine while escalating harder windows to the LLM-powered Trade Engine. Our contributions are as follows.

\begin{itemize}
    \item We cast missed-trade identification as \textbf{event extraction (EE) over multi-party dialogue}, a setting in which arguments are unanchored, spread across non-adjacent turns, and interleaved with concurrent RFQ events. We show that direct off-the-shelf LLM prompting over full chatrooms performs poorly on this task.

    \item We show that \textbf{per-event RFQ windowing combined with compact specialized classifiers as inference-time scaffolds substantially improves GPT-4o over full-chatroom CoT prompting} for multi-party dialogue EE. The resulting pipeline, \textbf{FinDialogLens}, uses these classifiers as inference-time scaffolds for LLM extraction and achieves 92.1\% and 94.3\% accuracy for final price and trade outcome extraction.

    \item We propose a \textbf{difficulty-aware router} supervised by the per-event correctness of a rule-based weak engine. For final price, routing around 15\% of RFQs to the LLM-powered Trade Engine (an 85\% reduction in LLM calls) recovers half of the accuracy gap between the weak engine and FinDialogLens (GPT-4o), saving over \$300/day at our 70{,}000-RFQ/day scale.
\end{itemize}

\section{Related Work}
\label{sec:related}

\paragraph{Event extraction and financial NLP.}
Event extraction (EE) has progressed from sentence-level formulations on ACE~\citep{doddington-etal-2004-automatic} to document-level argument linking~\citep{ebner-etal-2020-multi, li-etal-2021-document, zheng-etal-2019-doc2edag} and generative or instruction-tuned methods~\citep{lu2021text2event, hsu2022degree, wang2023instructuie, sainz2024gollie}. Financial NLP~\citep{li2023survey, nie2024finsurvey, chen2024critical} applies structured information extraction (IE) primarily to news and filings (e.g.\ Doc2EDAG~\citep{zheng-etal-2019-doc2edag}, FinRED~\citep{sharma2022finred}); conversational financial data receives little attention. Dialogue disentanglement~\citep{kummerfeld-etal-2019-large, Li:2024} handles multi-party input but targets thread separation rather than event extraction. Our setting requires four structural properties simultaneously: multi-party input, arguments that span multiple turns, explicit per-event triggers, and structured event records as output.

\begin{table}[t]
\centering
\small
\setlength{\tabcolsep}{4pt}
\begin{tabular}{lcccc}
\toprule
Method family & Multi- & Cross- & Trigger- & Event \\
              & party  & turn   & anchored & record \\
\midrule
Sentence EE       & $\times$    & $\times$    & $\checkmark$ & $\checkmark$ \\
Doc-level EE      & $\times$    & $\checkmark$ & $\times$    & $\checkmark$ \\
Dialogue disent.  & $\checkmark$ & $\checkmark$ & --          & $\times$    \\
\midrule
\textbf{Ours}     & $\checkmark$ & $\checkmark$ & $\checkmark$ & $\checkmark$ \\
\bottomrule
\end{tabular}
\caption{Structural axes along which existing EE method families differ from our application setting.}
\label{tab:ee-comparison}
\end{table}

\section{Methodology}

\subsection{Problem Setting}

A chatroom $C = (m_1,\ldots,m_N)$ is a time-ordered sequence of messages, each tagged with an author identifier. We model each RFQ as a typed event anchored by a single \emph{RFQ-trigger message} and associated with two argument roles: \emph{final price} and \emph{trade outcome}. Throughout, an RFQ-trigger message refers to the message that initiates an RFQ. The goal is to recover the set of $K$ RFQ events present in $C$,
\begin{equation}
    E = \{e_j\}_{j=1}^{K}, \quad e_j = (m^{\text{trig}}_j,\, p_j,\, o_j),
\end{equation}
where $m^{\text{trig}}_j \in C$ is the RFQ-trigger message that anchors the event, $p_j$ is the final price, represented as a string lifted from the dialogue or \textsc{None} if no price applies, and $o_j \in \{\text{Hit}, \text{Miss}, \text{Unknown}\}$ is the trade outcome.\footnote{\textsc{Hit}: the client agreed to trade; \textsc{Miss}: the client explicitly declined (e.g.\ ``away''); \textsc{Unknown}: no client response was observed within the context window. \textsc{Unknown} is operationally distinct from \textsc{Miss} because sales follow-up decisions differ in the two cases.} Evidence for the two argument roles may appear well after $m^{\text{trig}}_j$ and can be interleaved with messages from other concurrent RFQ events.

\subsection{Message-Level Module}

The Message-Level Module produces message-level metadata that serves as inference-time scaffolding for RFQ windowing and argument role filling. Inspired by recent work~\citep{yu2023open} showing that fine-tuning compact models like RoBERTa with sufficient data can outperform larger LLM-based models in specialized tasks, we fine-tune three RoBERTa-base models on in-domain data. The \emph{RFQ Identifier} is a binary trigger detector that predicts whether each $m_i \in C$ initiates an RFQ event. The \emph{Trade Outcome Classifier} assigns each message a label in $\{\text{Hit}, \text{Miss}, \text{None}\}$, highlighting messages that may express trade outcomes. A \emph{Price NER Model} extracts price entities from each message. The resulting metadata annotations are attached to each $m_i$ and consumed by the RFQ-Level Module and the Trade Engine.\footnote{Training details are reported in Appendix~\ref{appendix:message-level module}.}

\subsection{RFQ-Level Module}

A typical daily chatroom $C$ may contain hundreds of messages and numerous RFQs, with multiple clients and sales representatives negotiating prices for different products while also exchanging unrelated market updates and informal chatter. Processing the full chatroom context is therefore both inefficient and potentially unreliable, as long-context LLMs may underutilize relevant evidence when it is buried within lengthy inputs, a phenomenon commonly described as the ``lost-in-the-middle'' effect~\citep{liu2024lost}. To address this, the RFQ-Level Module constructs an \emph{RFQ window}
for each detected RFQ-trigger message. Each RFQ window is a temporally bounded, per-event context anchored at the trigger and passed to the Trade Engine for argument role filling. We denote the window for trigger $m^{\text{trig}}_j$ as $w_j$.

For each trigger $m^{\text{trig}}_j$ flagged by the RFQ Identifier, we collect the trigger message and its subsequent messages in $C$ until either a 20-minute time gap or a 25-message cap is reached.\footnote{Thresholds were chosen on the evaluation split; the resulting window covers the gold final price and trade outcome for over 96\% of RFQ events.} The message-level metadata is interleaved with the raw text inside $w_j$ using special tokens (Figure~\ref{fig:input_context}). The module outputs the set of RFQ windows $W = \{w_j\}$, one for each trigger detected by the RFQ Identifier.

\subsection{Trade Engine}

Given a per-event RFQ window $w_j \in W$ anchored at trigger message $m^{\text{trig}}_j$, the Trade Engine predicts the remaining argument roles: the final price $p_j$ and trade outcome $o_j$. The full event record is then assembled as $e_j = (m^{\text{trig}}_j, p_j, o_j)$.

The Trade Engine is LLM-powered. It prompts an LLM with the metadata-augmented RFQ window $w_j$ under a unified template (Figure~\ref{fig:trade_engine_prompt}), then parses the structured response into $(p_j, o_j)$. The LLM is instructed to identify the final price when applicable and determine the trade outcome from the dialogue evidence. We instantiate the Trade Engine with two LLM families:

\begin{enumerate}
    \item \textbf{Proprietary}: GPT-4o.
    \item \textbf{Open-source}: Mistral-7B-Instruct~\cite{jiang2023mistral7b}, Llama-3-8B-Instruct~\cite{grattafiori2024llama3herdmodels}, Phi-3-mini-4k-instruct~\cite{abdin2024phi3technicalreporthighly}, and Flan-T5-XL~\cite{chung2022scalinginstructionfinetunedlanguagemodels}.
\end{enumerate}
Throughout, ``FinDialogLens ($M$)'' denotes the full pipeline using model $M$ as the LLM-powered Trade Engine.

\begin{figure}[t]
    \centering
    \includegraphics[width=0.4\textwidth]{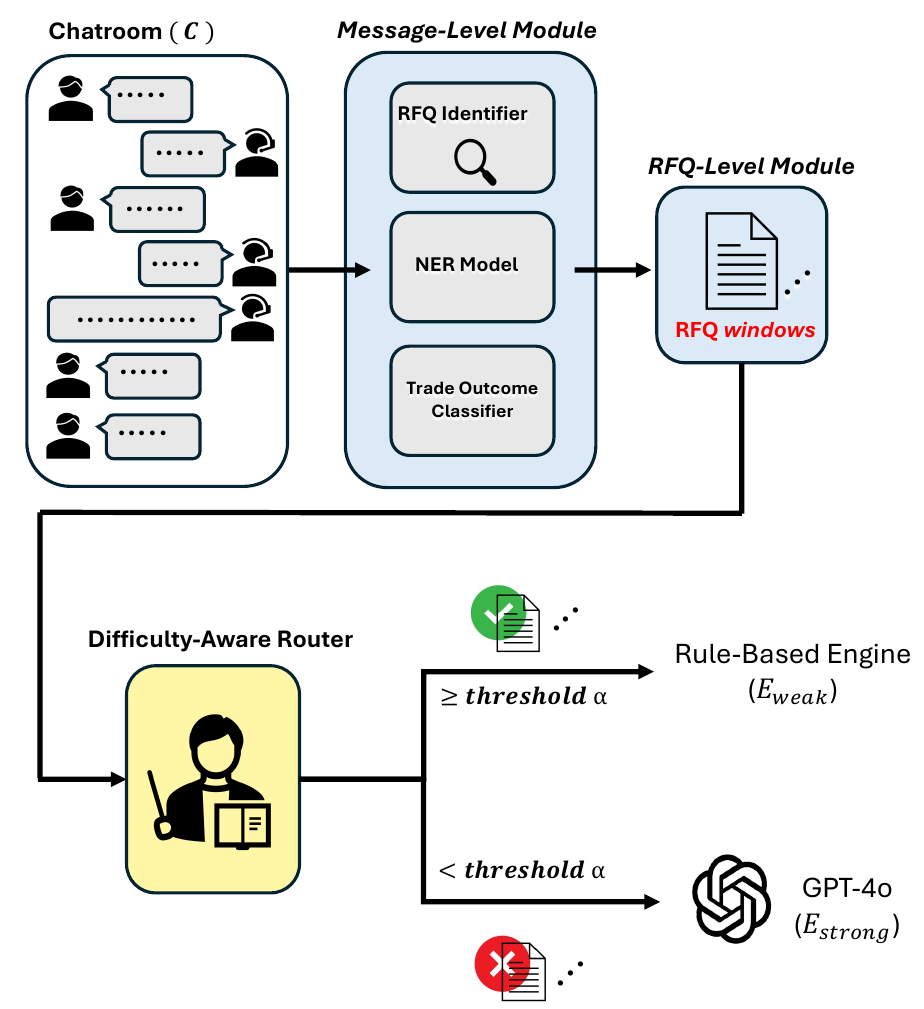}
    \caption{Difficulty-Aware Router. For each RFQ event, the router assigns easier cases to rule-based engine and escalates harder cases to the LLM-powered engine.}
    \label{fig:framework-left}
\end{figure}

\subsection{Difficulty-Aware Router}

Although the LLM-powered Trade Engine achieves strong performance,
using it for every RFQ window incurs avoidable cost and latency. Some
windows exhibit empirical structural patterns, such as nearby evidence,
predictable reply behavior, or explicit product and outcome references,
that a rule-based weak engine $E_{\text{weak}}$ can exploit reliably.
As shown in Figure~\ref{fig:framework-left}, we therefore introduce a \emph{difficulty-aware router} that sends
lower-ambiguity RFQ windows to $E_{\text{weak}}$ and reserves the
LLM-powered strong engine $E_{\text{strong}}$ for more ambiguous cases.

We train the router to estimate the reliability of the weak engine,
$P_\theta(\text{Correct}\mid w)$, using labels obtained by running the
weak engine on labeled RFQ windows. At inference time, a threshold
$\alpha$ controls the cost--accuracy trade-off:
\begin{equation}
R^\alpha(w) =
\begin{cases}
E_{\text{weak}}, & P_\theta(\text{Correct}\mid w) \geq \alpha, \\
E_{\text{strong}}, & \text{otherwise.}
\end{cases}
\end{equation}
We instantiate $P_\theta$ with RoBERTa-base, Mistral-7B-Instruct, and
Flan-T5-XL.
\begin{table*}[t]
\centering
\resizebox{0.9\textwidth}{!}{
\newcommand{\std}[1]{{\scriptstyle \pm #1}}

\begin{tabular}{c|c|cc|c|c|cc}
\toprule
\multirow{2}{*}{Method} 
    & \multirow{2}{*}{w/ Metadata} 
    & \multicolumn{2}{c|}{Input Context} 
    & \multirow{2}{*}{Trade Engine} 
    & \multicolumn{1}{c|}{Final Price} 
    & \multicolumn{2}{c}{Trade Outcome} \\
\cmidrule(lr){3-4} \cmidrule(lr){6-6} \cmidrule(lr){7-8}
& & Whole Day & RFQ Windows & & Acc & Acc & F1 \\
\midrule
Rule-based Engine
    & $\checkmark$ & $\checkmark$ & & Rule-based
    & 77.1
    & 90.3
    & 90.7 \\
\midrule
\multirow{4}{*}{CoT Prompting} 
    &  & $\checkmark$ &  & GPT-4o (Zero-shot) & 60.9 & 65.5 & 77.1 \\
    &  & $\checkmark$ &  & GPT-4o (Few-shot)  & 64.4 & 70.5 & 80.5 \\
    & $\checkmark$ & $\checkmark$ &  & GPT-4o (Zero-shot)  & 73.9 & 85.9 & 91.5\\
    & $\checkmark$ & $\checkmark$ &  & GPT-4o (Few-shot)  & 79.2 & 87.0 & 92.3\\
\midrule
\multirow{5}{*}{FinDialogLens} 
    & $\checkmark$ &  & $\checkmark$ & GPT-4o & \textbf{92.1} & \textbf{94.3} & 94.5 \\
    & $\checkmark$ &  & $\checkmark$ & Mistral-7B-Instruct (Fine-tuned) & 89.0 & 92.7 & \textbf{94.6} \\
    & $\checkmark$ &  & $\checkmark$ & Llama-3-8B-Instruct (Fine-tuned) & 87.1 & 90.3 & 92.8 \\
    & $\checkmark$ &  & $\checkmark$ & Phi-3-mini-4k-instruct (Fine-tuned) & 87.8 & 91.2 & 92.9 \\
    & $\checkmark$ &  & $\checkmark$ & Flan-T5-XL (Fine-tuned) & 91.7 & 92.8 & 94.4 \\
\bottomrule
\end{tabular}
}
\caption{
Comparison of final price and trade outcome extraction performance across the Rule-based Engine, full-chatroom CoT prompting baselines, and FinDialogLens.
``w/ Metadata'' indicates whether the input context is augmented with predicted RFQ-trigger, price, and trade-outcome tags from the Message-Level Module. 
``Input Context'' specifies whether the model consumes the full daily chatroom or per-event RFQ windows produced by the RFQ-Level Module.
}
\label{table1}
\end{table*}

\section{Experiments}

\subsection{Experiment Setup} 

\subsubsection{Data Overview}

We annotate 47{,}362 messages from two years of financial conversations (2023--2024), covering 8{,}343 RFQs in total. The data is partitioned into three non-overlapping splits: 37{,}245 messages / 7{,}082 RFQs for training, 1{,}147 messages / 295 RFQs for evaluation, and 8{,}970 messages / 966 RFQs for testing. The splits are time-based rather than uniformly sampled: we hold out a contiguous June--July 2024 window, the most recent period in our annotated data, as the test set. The training and evaluation sets are drawn from the remaining 2023--2024 data, with no RFQ or message overlap across splits.

This setup mimics the production setting, where models are trained on historical labeled data and evaluated on new incoming conversations. The training split is used to train the Message-Level Module, Trade Engine, and difficulty-aware router, while the evaluation split is used for prompt and hyperparameter selection. All reported results are on the test split. To protect sensitive financial communications, strict privacy protocols are enforced to prevent data leakage or unauthorized access.\footnote{Further details on data privacy, annotation guidelines, and training details are provided in Appendix~\ref{appendix:data} and Appendix~\ref{appendix: training details}.}

\subsubsection{Baselines}

We compare FinDialogLens against both non-LLM and LLM-based baselines. 
As a non-LLM baseline, we implement a Rule-based Engine that uses the same message-level metadata produced by the Message-Level Module and links predicted prices and trade outcomes to RFQ events through predefined heuristics based on conversation characteristics.
As LLM-based baselines, we evaluate end-to-end Chain-of-Thought (CoT) prompting~\citep{3600270.3602070} with GPT-4o (\texttt{gpt-4o-2024-05-13}). 
We run CoT prompting on the full daily chatroom under four configurations: 
$\{\text{zero-shot}, \text{few-shot (3 examples)}\} \times \{\text{with metadata}, \text{without metadata}\}$. 
In the metadata setting, the full chatroom is augmented with the Message-Level Module's predicted RFQ-trigger, price, and trade-outcome tags using the same special tokens as FinDialogLens (Figure~\ref{fig:input_context}). 
The CoT prompt templates are shown in Figure~\ref{fig:cot_prompt}.

\subsubsection{Evaluation Protocol and Metrics}
\label{sec: exp_metrics}

We evaluate FinDialogLens with a trigger-anchored event extraction protocol. The RFQ-trigger message defines the identity of an RFQ event, while final price and trade outcome are arguments whose correctness is defined relative to that anchor. At inference time, FinDialogLens runs end-to-end from raw chatroom messages: it detects RFQ triggers and metadata, constructs one RFQ window for each detected trigger, and predicts the final-price and trade-outcome arguments. Gold triggers are
not provided as input.

For scoring, predicted RFQ events are aligned to reference events by their trigger message, and argument accuracy is computed on the aligned events. RFQ-trigger detection is evaluated separately through the RFQ Identifier in the Message-Level Module, with precision, recall, and F1 reported in Table~\ref{table:Message-Level-models}; this separates event-anchor discovery from cross-turn argument linking. Final price is evaluated by exact-match accuracy with no normalization or post-processing; trade outcome is evaluated as a three-way classification over Hit, Miss, and Unknown using accuracy and F1.

To assess the difficulty-aware router, we follow RouteLLM~\citep{ong2025routellmlearningroutellms} and report Performance Gap Recovery (PGR) and Call-Performance Threshold (CPT):
\begin{equation}
PGR(R^\alpha) = \frac{r(R^\alpha) - r(E_\text{weak})}{r(E_\text{strong}) - r(E_\text{weak})},
\end{equation}
where $E_\text{weak}$ is the rule-based weak engine, $E_\text{strong}$ is the LLM-powered Trade Engine with GPT-4o, and $r(\cdot)$ is response quality measured by accuracy. CPT($x\%$) is the minimum percentage of RFQ windows routed to $E_\text{strong}$ needed to achieve PGR$\,=\,x\%$.

\subsection{Results}

We evaluate FinDialogLens along three dimensions aligned with our main claims: 
(1) whether message-level metadata and per-event RFQ windowing improve LLM-based role filling over full-chatroom CoT prompting; 
(2) whether fine-tuned open-source LLMs can approach proprietary LLM performance within the FinDialogLens pipeline; and 
(3) whether the difficulty-aware router can reduce LLM usage while preserving extraction accuracy. 
As a prerequisite, we verify that the Message-Level Module provides reliable metadata scaffolds for downstream windowing and role filling. The three specialized RoBERTa-base classifiers achieve F1 scores of $0.97$, $0.95$, and $0.93$ for the RFQ Identifier, Trade Outcome Classifier, and Price NER model, respectively; per-model precision and recall are reported in Table~\ref{table:Message-Level-models}.

\begin{figure*}[t]
    \centering
    \includegraphics[width=0.7\textwidth]{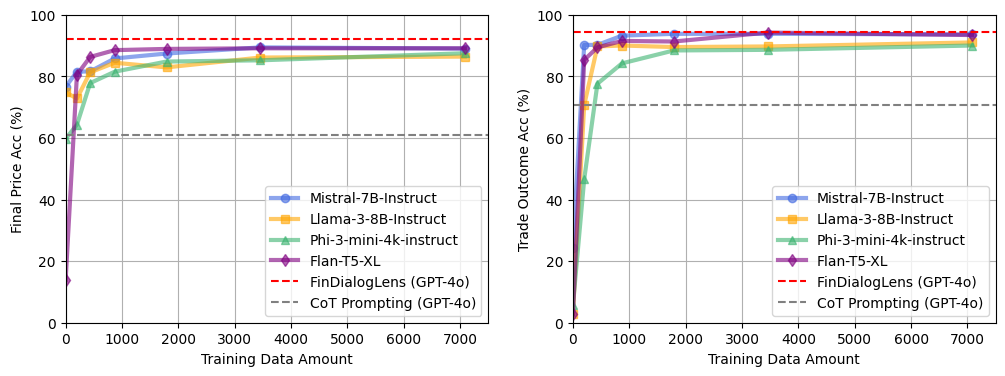}

 \caption{Accuracy of fine-tuned open-source LLMs on final price and trade outcome extraction as training-set size varies. Dashed lines mark FinDialogLens (GPT-4o) and zero-shot CoT Prompting (GPT-4o) baseline.}
    \label{fig:open-source}
\end{figure*}

\begin{table}[t]
    \centering
    \resizebox{0.9\linewidth}{!}{%
    \begin{tabular}{lccc}
    \toprule
    Model                    & Precision & Recall & F1    \\
    \midrule
    RFQ Identifier           & 0.97      & 0.97   & 0.97    \\
    Trade Outcome Classifier & 0.96      & 0.93   & 0.95    \\
    Price NER Model& 0.91      & 0.95   & 0.93    \\
    \bottomrule
    \end{tabular}
    }
    \caption{Performance of the Message-Level Module, including the RFQ Identifier, Trade Outcome Classifier, and Price NER Model.}
    \label{table:Message-Level-models}
\end{table}

\begin{table*}[t]
\centering
\resizebox{0.6\textwidth}{!}{
\begin{tabular}{lcc|cc}
\toprule
\multirow{2}{*}{Router Type} 
    & \multicolumn{2}{c|}{Final Price} 
    & \multicolumn{2}{c}{Trade Outcome} \\
\cmidrule(lr){2-3} \cmidrule(lr){4-5}
     & CPT (50\%) & CPT (80\%) 
     & CPT (50\%) & CPT (80\%) \\
\midrule
Random             & 49.2\% & 81.3\%  & 52.1\% & 75.8\% \\
RoBERTa-base       & \textbf{14.9\%} & 32.4\%  & 29.5\% & 62.6\% \\
Mistral-7B-Instruct         & 16.4\% & 31.8\%  & 26.7\% & 36.3\% \\
Flan-T5-XL         & 16.1\% & \textbf{29.9\%} & \textbf{25.1\%} & \textbf{32.8\%} \\
\bottomrule
\end{tabular}
}
\caption{
Call-Performance Threshold (CPT) for different router types. CPT is the fraction of RFQ windows routed to the LLM-powered Trade Engine required to recover a target percentage of the gap between the rule-based weak engine and FinDialogLens (GPT-4o). Bold indicates the best (lowest) value per column.
}
\label{tab:cpt}
\end{table*}

\subsubsection{Effect of Message-Level Metadata and Per-Event Windowing}
\label{sec:research question 1}

Table~\ref{table1} compares FinDialogLens with the Rule-based Engine and full-chatroom CoT-prompting baselines. The ``w/ Metadata'' column denotes whether the input is augmented with predicted RFQ-trigger, price, and trade-outcome tags produced by the Message-Level Module. The CoT-prompting baselines apply GPT-4o to the full daily chatroom, whereas FinDialogLens first segments the chatroom into RFQ-specific windows and then performs role filling within each window. The main table reports one run for each method: GPT-4o-based methods use decoding with temperature set to 0, and fine-tuned open-source LLMs use greedy decoding. Because hosted API models can still exhibit residual nondeterminism even with temperature set to 0, Appendix~\ref{app:gpt4o-multirun} reports five-run mean and standard deviation for GPT-4o-based methods.

Message-level metadata substantially improves both the Rule-Based Engine and full-chatroom CoT prompting methods. For the Rule-Based Engine, message-level metadata enables final price and trade outcome accuracies of 77.1\% and 90.3\%, respectively. This improvement is largely due to the strong performance of the Message-Level Module, which helps the Rule-Based Engine link the correct final price and trade outcome to each detected RFQ trigger. However, the rule-based approach still struggles when multiple RFQs occur within a very short time span.

In the few-shot CoT setting, adding metadata increases final price accuracy from 64.4\% to 79.2\% and trade outcome accuracy from 70.5\% to 87.0\%, corresponding to gains of 14.8 and 16.5 points, respectively. Similar improvements appear in the zero-shot setting, where metadata improves final price accuracy by 13.0 points and trade outcome accuracy by 20.4 points. These results show that compact message-level classifiers provide useful domain-specific signals for LLM-based extraction from noisy financial dialogue.

Per-event RFQ windowing provides a second, complementary gain. Compared with the strongest metadata-augmented CoT baseline, FinDialogLens (GPT-4o) improves final price accuracy from 79.2\% to 92.1\% and trade outcome accuracy from 87.0\% to 94.3\%, yielding gains of 12.9 and 7.3 points, respectively. This suggests that isolating each RFQ into a focused context window reduces cross-event interference and makes argument role filling easier for the LLM.

Overall, message-level metadata and per-event windowing jointly drive FinDialogLens's improvement. Metadata provides compact domain-specific signals, while per-event windowing focuses the LLM on the relevant RFQ context for role filling. Together, these components allow FinDialogLens to achieve the best overall performance among all compared methods, with 92.1\% final price accuracy and 94.3\% trade outcome accuracy.

\subsubsection{Open-source vs.\ Proprietary LLMs}
\label{sec:research question 2}

Figure~\ref{fig:open-source} shows that fine-tuned open-source LLMs in FinDialogLens improve rapidly on both final price and trade outcome extraction as the amount of in-domain training data increases, with performance plateauing after about 1{,}000 examples. With the full training set, all four open-source LLMs outperform GPT-4o CoT prompting baselines in Table~\ref{table1}. For instance, Flan-T5-XL achieves 91.7\% final price accuracy and 92.8\% trade outcome accuracy. These results show that modest in-domain supervision enables open-source LLMs with as few as 3B parameters to approach proprietary-model performance within the FinDialogLens pipeline.

\begin{figure}[t]
    \centering
    \includegraphics[width=0.4\textwidth]{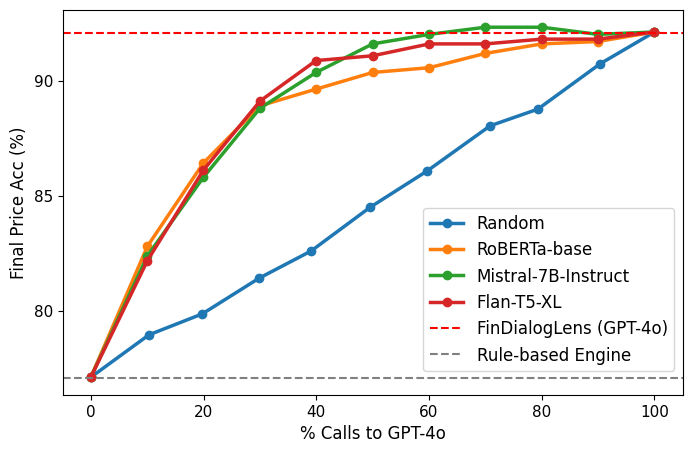}
    \caption{Performance of routers under various budget constraints ($\alpha$).}
    \label{fig:budget}
\end{figure}

\subsubsection{Difficulty-Aware Router}
\label{sec:research question 3}

The difficulty-aware router assigns each RFQ window to either the rule-based weak engine or the LLM-powered Trade Engine. The weak engine achieves 77.1\% final price accuracy and 90.3\% trade outcome accuracy on the test set, compared with 92.1\% and 94.3\% for FinDialogLens (GPT-4o), respectively. We evaluate routing by how efficiently it recovers this performance gap while reducing LLM calls, using both CPT and threshold-based analyses to measure the trade-off between accuracy and LLM usage.

Table~\ref{tab:cpt} reports the Call-Performance Thresholds (CPT) for four routing strategies. Lower CPT indicates greater efficiency: fewer RFQ windows need to be routed to the LLM-powered Trade Engine to achieve a target PGR. Fine-tuned routers substantially outperform the Random baseline. For final price, RoBERTa-base reaches 50\% PGR while routing only 14.9\% of RFQ windows to the LLM-powered Trade Engine, and Flan-T5-XL reaches 80\% PGR with 29.9\%. For trade outcome, Flan-T5-XL requires only 25.1\% and 32.8\% of RFQ windows to reach 50\% and 80\% PGR, respectively.

Figure~\ref{fig:budget} shows the effect of varying $\alpha$, which adjusts the fraction of RFQ windows routed to the LLM-powered engine. Fine-tuned routers outperform the Random router in final price accuracy as the LLM call fraction decreases. Notably, they achieve performance close to running the LLM-powered engine on every RFQ window while routing only 40\% of RFQ windows to it, indicating their effectiveness in assigning easier RFQ windows to the rule-based engine. Among the fine-tuned models, performance is similar, with no single model consistently superior.

Detailed routing-quality, cost, and latency analyses are in Appendix~\ref{appndix:cost}. In our production setting of over 70{,}000 RFQs/day, routing 14.9\% of RFQ windows to the LLM-powered Trade Engine for final price extraction recovers half of the accuracy gap between the weak engine and FinDialogLens (GPT-4o), saving over \$300/day in API spend.

\section{Conclusion}

We present \textbf{FinDialogLens}, a hybrid LLM pipeline for event extraction over multi-party dialogue, applied to missed-trade identification in financial chatrooms. Compact message-level classifiers provide inference-time metadata scaffolds, the RFQ-Level Module segments per-event RFQ windows, and the LLM-powered Trade Engine predicts the final price and trade outcome for each recovered RFQ. With GPT-4o, FinDialogLens reaches 92.1\% and 94.3\% accuracy on final price and trade outcome, outperforming full-chatroom CoT prompting; fine-tuned open-source LLMs with as few as 3B parameters achieve comparable performance with modest in-domain data. A difficulty-aware router further reduces cost by assigning easier RFQs to a rule-based weak engine and escalating harder RFQs to the LLM-powered Trade Engine, cutting LLM calls by 85\% on final price while recovering half of the accuracy gap to FinDialogLens (GPT-4o), saving over \$300/day at our 70{,}000-RFQ/day scale.

\section*{Limitations}

FinDialogLens provides a reusable framework for missed-trade identification in financial multi-party dialogue, a setting that has received limited attention in prior event extraction research. While the framework is designed to generalize beyond a single deployment environment, several components require adaptation before being applied to new institutions, asset classes, or communication settings.

First, FinDialogLens depends on high-performing compact models to generate message-level metadata, including RFQ-trigger labels, price mentions, and trade outcome cues. These signals are central to the pipeline: they guide RFQ window construction and provide inference-time scaffolding that helps the downstream LLM focus on relevant evidence. However, training reliable compact models requires labeled in-domain data, and their utility may decrease when annotations are scarce or when communication styles and product conventions shift. Second, the RFQ-Level Module currently constructs per-event RFQ windows using fixed temporal and message-count thresholds. Although these thresholds provide strong coverage in our setting, negotiation dynamics can vary substantially across desks, clients, and asset classes; some RFQs resolve immediately, while others unfold over longer intervals with clarification, requoting, or follow-up. Adaptive boundary detection conditioned on dialogue structure, participant behavior, and product-specific patterns could improve portability. Third, our difficulty-aware router is tailored to our deployment setting, where the primary decision is whether to use a low-cost rule-based engine or escalate to an LLM-powered Trade Engine. This binary design captures the cost–accuracy trade-off most relevant to our use case, but other deployments may require multi-way, cost-aware, or task-specific routing across multiple model tiers, specialist engines, latency constraints, or risk-based escalation policies.

Finally, due to client and regulatory confidentiality, the underlying conversations cannot be publicly released, limiting direct external replication on the original corpus. To partially mitigate this limitation, we provide annotation guidelines, dataset statistics, model architectures, training configurations, hyperparameters, and evaluation protocols to support reproduction on comparable proprietary financial dialogue datasets.

\section*{Ethics Statement}

This research is driven by the goal of enhancing operational efficiency in financial markets through the automated analysis of unstructured communication channels, such as chatrooms and instant messaging platforms. The proposed framework, FinDialogLens, is designed to support financial professionals in identifying missed trading opportunities and improving decision-making, while complementing, rather than replacing, human expertise and judgment.

The financial data and conversational content analyzed in this study are highly sensitive and managed in accordance with strict industry governance and privacy protocols. All experiments were conducted using anonymized, internally governed datasets, with user identities replaced by generic identifiers to ensure privacy. No personal or client-identifying information was collected, stored, or utilized at any stage. Access to both data and model outputs is strictly restricted to authorized personnel, and all data handling complies with relevant regulatory and organizational standards.

The sample dialogues presented in this paper do not represent the opinions or positions of the authors or their organizations. We are committed to maintaining the highest standards of data privacy, security, and research integrity throughout this work.

\section*{Disclaimer}
This paper was prepared for informational purposes in part by the Machine Learning Center of Excellence group of JPMorgan Chase \& Co. and its affiliates (``JP Morgan'') and is not a product of the Research Department of JP Morgan. JP Morgan makes no representation and warranty whatsoever and disclaims all liability, for the completeness, accuracy or reliability of the information contained herein. This document is not intended as investment research or investment advice, or a recommendation, offer or solicitation for the purchase or sale of any security, financial instrument, financial product or service, or to be used in any way for evaluating the merits of participating in any transaction, and shall not constitute a solicitation under any jurisdiction or to any person, if such solicitation under such jurisdiction or to such person would be unlawful.

\bibliography{custom}

\appendix

\section{Data Privacy, Annotation Guidelines, and Statistics}
\label{appendix:data}
\paragraph{Privacy:}To ensure the privacy and confidentiality of conversations used in our framework, we implemented dedicated anonymization measures. Specifically, user names within the conversations are replaced with generic identifiers such as “Client\_” and “Sales\_,” effectively safeguarding individual identities.

\paragraph{Annotation Guidelines:}
To ensure the accuracy and reliability of our human-labeled data, we adopted a rigorous annotation protocol involving two professional annotators, each of whom independently reviewed every message and Request for Quote (RFQ). For message-level annotation, both annotators labeled the RFQ-triggering message, trade outcome classification, and price information associated with each message. For RFQ-level annotation, each annotator independently assigned the final price and trading outcome for each RFQ, with outcomes categorized as Hit, Miss, or Unknown.

To maintain annotation quality, we excluded cases falling into either of the following categories:
(1) \textit{Annotator-disagreement cases}: cases in which both annotators were able to assign labels based on the chat context, but their labels differed.
(2) \textit{Unresolvable cases}: cases in which the gold final price or trade outcome could not be determined from the chatroom alone because the negotiation moved to other communication channels, such as phone or email.

The fraction of cases discarded due to direct annotator disagreement was very small, accounting for less than 1\% of the data. Because our annotators are professionals with domain experience, most labels were consistent across annotators. Separately, approximately 1\% of RFQs were marked as unresolvable because the raw chatroom data did not contain sufficient information to determine the gold final price or trade outcome.

We provide several examples of these unresolvable cases below. In these examples, XYZ and ABC are placeholders used to mask product or security identifiers for confidentiality reasons.

\paragraph{Examples of Unresolvable Cases.}
Below are examples of RFQs that were marked as unresolvable. In these cases, the final price or trade outcome cannot be determined from the chat text alone because the negotiation moved to another communication channel. Therefore, we exclude these cases as out of scope.

\begin{quote}
\textbf{Phone}\\
\texttt{<user\_1>: Could you please price XYZ?}\\
\texttt{<user\_2>: sure}\\
\texttt{<user\_2>: -16.5 / -15}\\
\texttt{<user\_1>: calling you}\\
\texttt{<user\_2>: ok}
\end{quote}

\begin{quote}
\textbf{Email}\\
\texttt{<user\_1>: Please price XYZ}\\
\texttt{<user\_2>: can u send an email}\\
\texttt{<user\_2>: resending email confirm}\\
\texttt{<user\_1>: email sent}
\end{quote}

\paragraph{Statistics of Challenging Interleaved RFQs.}
To quantify the prevalence of challenging interleaved RFQs in the test set, we categorize each test RFQ according to the number of gold RFQ-trigger messages appearing within its RFQ window. In total, 368 out of 966 RFQs, corresponding to 38.1\% of the test set, occur in windows containing two or more gold RFQ-trigger messages. These cases represent the type of interleaved RFQ scenarios illustrated in Figure~\ref{fig:edge_case}, where multiple RFQs appear in close proximity and a model may incorrectly associate a price or trading outcome with the wrong RFQ.

\begin{table}[t]
\centering
\resizebox{\columnwidth}{!}{%
\begin{tabular}{lcc}
\hline
\textbf{Difficulty} & \textbf{Count} & \textbf{Percentage} \\
\hline
Easy: 1 RFQ in window & 598 & 61.9\% \\
Hard: $\geq$2 RFQs in window & 368 & 38.1\% \\
\hline
Total & 966 & 100.0\% \\
\hline
\end{tabular}%
}
\caption{Distribution of test RFQs by the number of gold RFQ-trigger messages in the RFQ window. Hard cases correspond to interleaved RFQ windows containing two or more RFQs.}
\label{tab:interleaved_rfq_statistics}
\end{table}

\section{Training Details of FinDialogLens}
\label{appendix: training details}
All experiments are conducted with a fixed random seed of 42 and, where applicable, a temperature setting of 0. Model training is performed on a machine equipped with 4 NVIDIA A10G GPUs.

\subsection{Message-Level Module}
\label{appendix:message-level module}
Inspired by recent work \citep{yu2023open} demonstrating that fine-tuning compact models such as RoBERTa with sufficient data can outperform larger models like GPT-4 in specialized tasks, we fine-tune three RoBERTa-base models within the Message-Level module to generate more sophisticated financial outputs for each message. Formally, for each message $m_i \in C$:

\begin{itemize}
    \item \textbf{RFQ Identifier}: Determines whether a message $m_i$ constitutes a Request for Quote (RFQ). We define a binary function:
    \begin{equation}
        y^{\mathrm{RFQ}}_i = f_{\mathrm{RFQ}}(m_i) \in \{\mathrm{True}, \mathrm{False}\}
    \end{equation}
    where $y^{\mathrm{RFQ}}_i = True$ indicates that $m_i$ is identified as an RFQ message.

\item \textbf{Trade Outcome Classifier}: Determines whether a message $m_i$ indicates a trade outcome as either ``Hit'', ``Miss'', or ``None.'' We define:
    \begin{equation}
        y^{\mathrm{trade}}_i = f_{\mathrm{trade}}(m_i) \in \{\mathrm{Hit}, \mathrm{Miss}, \mathrm{None}\}
    \end{equation}
    \item \textbf{Price NER Model}: Identifies and extracts the price entity from the message content. We define:
    \begin{equation}
        y^{\mathrm{price}}_i = f_{\mathrm{NER}}(m_i)
    \end{equation}
    where $y^{\mathrm{price}}_i$ is the list of extracted price values from $m_i$.
\end{itemize}

The predictions $y^{\mathrm{RFQ}}_i$, $y^{\mathrm{trade}}_i$, and $y^{\mathrm{price}}_i$ are stored as metadata for each message $m_i$ and are utilized in the subsequent module of our framework. The precision, recall, and F1 scores for these models are reported in Table~\ref{table:Message-Level-models}.

\subsection{Trade Engine}
\begin{figure*}[t]
    \centering
    \includegraphics[width=0.9\textwidth]{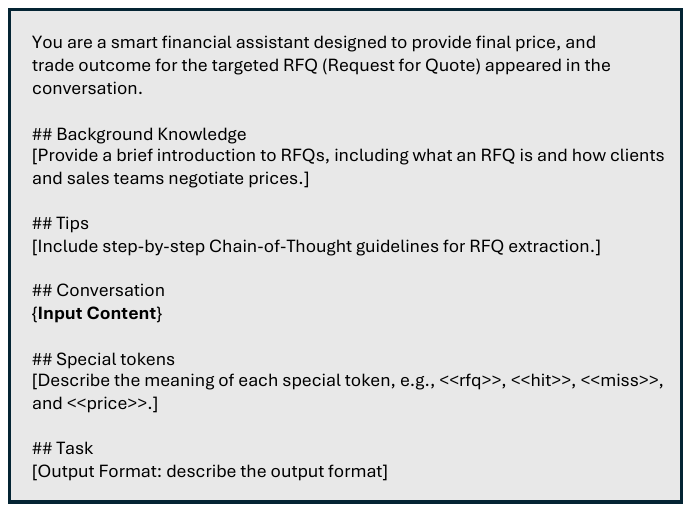}
    
    \caption{Prompt template for the Trade Engine within FinDialogLens.}
    \label{fig:trade_engine_prompt}
\end{figure*}

For the GPT-4o engine, we use ``gpt-4o-2024-05-13'' API version. We fine-tune four open-source LLMs as Trade Engines: (1) Mistral-7B-Instruct, (2) Llama-3-8B-Instruct, (3) Phi-3-mini-4k-instruct, and (4) Flan-T5-XL. The unified prompt template for all LLMs is shown in Figure~\ref{fig:trade_engine_prompt}. The training configuration is as follows:

\begin{enumerate}
    \item Batch\_size = 2
    \item Gradient\_accumulation\_steps = 16
    \item LoRAConfig = (lora\_alpha = 16, lora\_dropout = 0.1, r = 64)
    \item Optimizer: AdamW
    \item Epochs and learning rates (vary by training dataset size):
    
    \begin{center}
    \begin{tabular}{ccc}
    \toprule
    \#Train & \#Epoch & Learning rate \\
    \midrule
    200   & 10  & 5e-5 \\
    435   & 10  & 5e-5 \\
    870   & 10  & 5e-5 \\
    1790  & 20  & 5e-5 \\
    3455  & 20  & 5e-5 \\
    7082  & 20  & 5e-5 \\
    \bottomrule
    \end{tabular}
    \end{center}
\end{enumerate}

\subsection{Difficulty-Aware Router}
\label{appendix: type of routers}
In this section, we elaborate on the training details for three types of routers:

\begin{enumerate}
    \renewcommand{\labelenumi}{\alph{enumi})}
    \item \textbf{Encoder-only Classifier}: RoBERTa-base is fully fine-tuned for this classifier. A classification layer is applied to the \texttt{<s>} token embedding to determine the class label $ L $. The model is trained using cross-entropy loss, optimizing the parameters to accurately classify each input context window based on the provided labels.

    \item \textbf{Decoder-only Classifier}: Mistral-7B-Instruct is fine-tuned using a prompt template to generate a single token, either ``Yes'' or ``No.'' A ``Yes'' indicates that the rule-based engine is correct on the input context window, while a ``No'' indicates incorrectness. The probability of correctness is computed using a softmax over ``Yes'' and ``No,'' with ``Yes'' corresponding to $ P_\theta(\text{Correct} \mid w) $.

    \item \textbf{Encoder-Decoder Classifier}: Flan-T5-XL is fine-tuned for this classifier under the same training and inference pipeline as the Decoder-only Classifier, with both encoder and decoder used to process the input context window and emit a single token, either ``Yes'' or ``No''.
\end{enumerate}

\begin{figure}[t]
    \centering
    \includegraphics[width=0.45\textwidth]{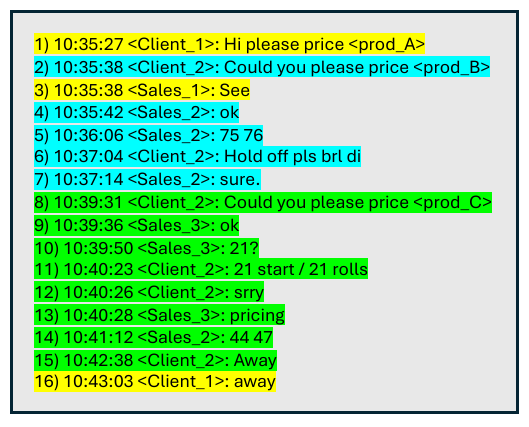}
    \caption{Example of an edge case in financial dialogue with three RFQs (messages 1, 2, and 8) occurring within a short time frame. Messages are highlighted in different colors to clarify which replies correspond to each RFQ event.}
    \label{fig:edge_case}
\end{figure}

\begin{figure}[t]
    \centering
    \includegraphics[width=0.45\textwidth]{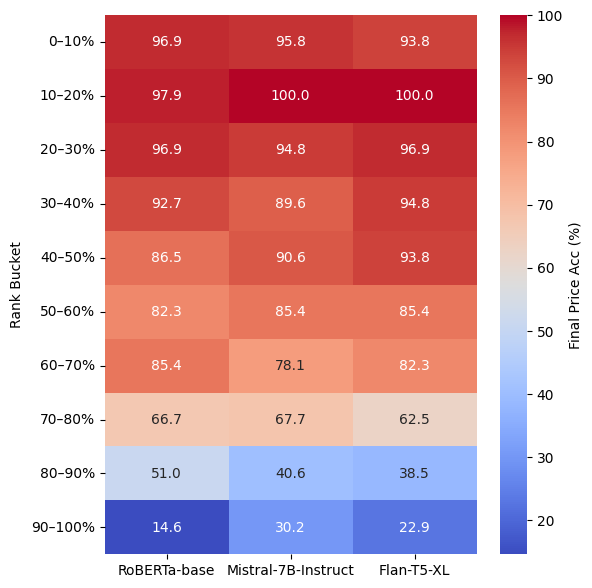}
    \caption{Accuracy of the rule-based engine on per-event RFQ windows grouped into deciles by router logit ranking.}
    \label{fig:quality}
\end{figure}

\begin{figure*}[t]
    \centering
    \includegraphics[width=0.9\textwidth]{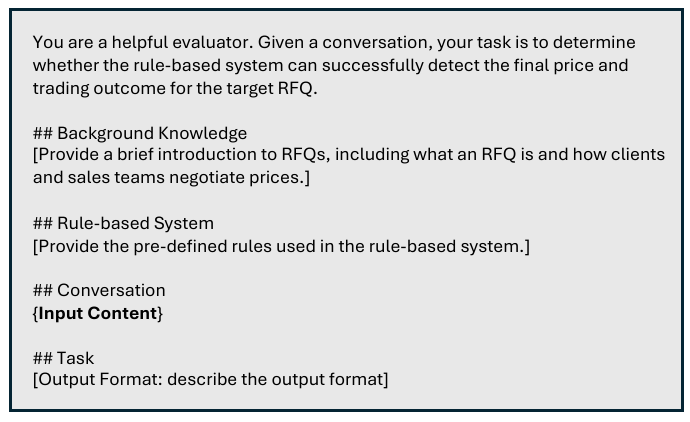}
    
    \caption{Prompt template for the difficulty-aware router.}
    \label{fig:router_prompt}
\end{figure*}

Below are the training configurations:

\textit{RoBERTa-base:}
\begin{enumerate}
    \item Epoch = 5
    \item Learning\_rate = 2e-5
    \item Batch\_size = 32
    \item Gradient\_accumulation\_steps = 8
    \item Optimizer: AdamW
\end{enumerate}
\textit{Mistral-7B-Instruct and Flan-T5-XL:} 
\begin{enumerate}
    \item Prompt template is shown in Figure~\ref{fig:router_prompt}
    \item Epoch = 5
    \item Learning\_rate = 1e-4
    \item Batch\_size = 4
    \item Gradient\_accumulation\_steps = 32
    \item LoRAConfig = (lora\_alpha = 16, lora\_dropout = 0.1, r = 64)
    \item Optimizer: AdamW
    
\end{enumerate}

\begin{table*}[t]
    \centering
    \resizebox{0.7\linewidth}{!}{%
    \begin{tabular}{lcc|cc}
    \toprule
    \multirow{2}{*}{Router Type} 
        & \multicolumn{2}{c|}{\bf Final Price} 
        & \multicolumn{2}{c}{\bf Trade Outcome} \\
    \cmidrule(lr){2-3} \cmidrule(lr){4-5}
         & CPT (50\%) & CPT (80\%) 
         & CPT (50\%) & CPT (80\%) \\
    \midrule
    Random             & \$2.66 & \$4.39  & \$2.81 & \$4.09 \\
    RoBERTa-base       & \textbf{\$0.80} & \$1.75  & \$1.59 & \$3.38 \\
    Mistral-7B-Instruct         & \$0.89 & \$1.72 & \$1.44 & \$1.96 \\
    Flan-T5-XL         & \$0.87 & \textbf{\$1.61} & \textbf{\$1.36} & \textbf{\$1.77} \\
    \bottomrule
    \end{tabular}
    }
    \caption{Cost analysis of different router types for final price and trade outcome under CPT (50\%) and CPT (80\%). Reported values are USD per 1{,}000 RFQs; processing 1{,}000 RFQs with GPT-4o costs approximately \$5.40.}
    \label{tab:cost}
\end{table*}

\begin{table*}[t]
    \begin{center}
    \resizebox{0.7\linewidth}{!}{%
    \begin{tabular}{lcc|cc}
    \toprule
    \multirow{2}{*}{Router Type} 
        & \multicolumn{2}{c|}{\textbf{Final Price}} 
        & \multicolumn{2}{c}{\textbf{Trade Outcome}} \\
    \cmidrule(lr){2-3} \cmidrule(lr){4-5}
         & CPT (50\%) & CPT (80\%) 
         & CPT (50\%) & CPT (80\%) \\
    \midrule
    Random             & 717.8 & 1186.2  & 760.1 & 1105.9\\
    RoBERTa-base       & \textbf{217.4} & 427.7  & 430.4 & 913.3\\
    Mistral-7B-Instruct         & 239.3 & 464.0 & 389.6 & 529.6 \\
    Flan-T5-XL         & 234.9 & \textbf{436.2} & \textbf{366.2} & \textbf{478.6} \\
    \bottomrule
    \end{tabular}
    }
    \end{center}
    \caption{Latency analysis of different router types for final price and trade outcome under CPT (50\%) and CPT (80\%). Reported values are total LLM API latency (seconds) for the full test set; processing the full test set with GPT-4o alone takes approximately 1{,}459 seconds.}
    \label{tab:latency}
\end{table*}

\section{Multi-Run Evaluation for GPT-4o-Based Methods}
\label{app:gpt4o-multirun}

For GPT-4o-based methods, we use temperature-0 decoding.
Although this setting is near-greedy, API-side nondeterminism
can still introduce small run-to-run variation. We therefore
run each GPT-4o-based setting five times and report the mean
and standard deviation in Table~\ref{tab:gpt4o-multirun}.

\begin{table*}[t]
\centering
\small
\setlength{\tabcolsep}{4.5pt}
\renewcommand{\arraystretch}{1.08}
\begin{tabular}{@{}l c cc l ccc@{}}
\toprule
\multirow{2}{*}{\textbf{Method}}
    & \multirow{2}{*}{\textbf{w/ Metadata}}
    & \multicolumn{2}{c}{\textbf{Input Context}}
    & \multirow{2}{*}{\textbf{Trade Engine}}
    & \textbf{Final Price}
    & \multicolumn{2}{c}{\textbf{Trade Outcome}} \\
\cmidrule(lr){3-4} \cmidrule(lr){6-6} \cmidrule(lr){7-8}
& & \textbf{Whole Day} & \textbf{RFQ Windows}
& & \textbf{Acc} & \textbf{Acc} & \textbf{F1} \\
\midrule

\multirow{4}{*}{CoT Prompting}
    &  & \checkmark &  & GPT-4o (Zero-shot)
    & $62.2_{\pm 0.95}$
    & $66.7_{\pm 0.94}$
    & $76.7_{\pm 0.37}$ \\

    &  & \checkmark &  & GPT-4o (Few-shot)
    & $66.1_{\pm 2.17}$
    & $67.5_{\pm 2.47}$
    & $77.2_{\pm 2.09}$ \\

    & \checkmark & \checkmark &  & GPT-4o (Zero-shot)
    & $76.4_{\pm 1.73}$
    & $84.4_{\pm 1.32}$
    & $89.7_{\pm 1.40}$ \\

    & \checkmark & \checkmark &  & GPT-4o (Few-shot)
    & $79.2_{\pm 0.59}$
    & $85.2_{\pm 0.81}$
    & $90.3_{\pm 0.93}$ \\

\midrule

FinDialogLens
    & \checkmark &  & \checkmark & GPT-4o
    & $\mathbf{91.8}_{\pm 0.42}$
    & $\mathbf{93.4}_{\pm 0.27}$
    & $\mathbf{93.5}_{\pm 0.38}$ \\

\bottomrule
\end{tabular}
\caption{Multi-run results for GPT-4o-based methods. We report
mean accuracy/F1 over five runs, with standard deviation in
subscript. All GPT-4o runs use temperature-0 decoding.}
\label{tab:gpt4o-multirun}
\end{table*}

\section{Error Analysis}
We conducted error analysis of FinDialogLens (GPT-4o) on the test data and found that most edge cases occurred when multiple RFQs were requested within a short time period. In these complex scenarios, the LLM occasionally failed to correctly identify which response message corresponded to which RFQ.

For example, in Figure~\ref{fig:edge_case}, the target RFQ is the first message of the conversation, where <Client\_1> requests a price for <prod\_A>. Shortly after, <Client\_2> submits a second RFQ for <prod\_B> (message 2), and both RFQs receive prompt replies from <Sales\_1> and <Sales\_2>, respectively. Notably, <Sales\_1> does not provide any price for the target RFQ, while <Sales\_2> offers a price of ``75 76'' in message 5, which corresponds to the second RFQ. Eventually, <Client\_1> replies ``away'' in message 16, indicating the target RFQ was a missed trade. In this scenario, the LLM correctly identifies the target RFQ as a missed trade based on message 16, but incorrectly links the price ``75 76'' from message 5 to the target RFQ. The correct final price for the target RFQ should be \textsc{None}.

\section{Ablation Study on Metadata Error Propagation}
\label{app:metadata_error_propagation}

To quantify the impact of metadata errors and the extent to which they propagate to the downstream RFQ-level prediction task, we conduct an additional ablation study. In this study, we replace the predicted message-level price and trade-outcome metadata with human-annotated gold labels, while keeping GPT-4o as the Trade Engine. This allows us to compare the original pipeline, which uses predicted metadata, against an oracle setting in which the downstream Trade Engine receives gold message-level metadata.

\begin{table*}[t]
\centering
\small
\setlength{\tabcolsep}{4.5pt}
\renewcommand{\arraystretch}{1.08}
\begin{tabular}{@{}lccc@{}}
\toprule
\multirow{2}{*}{\textbf{Method}}
    & \textbf{Final Price}
    & \multicolumn{2}{c}{\textbf{Trade Outcome}} \\
\cmidrule(lr){2-2} \cmidrule(lr){3-4}
    & \textbf{Acc}
    & \textbf{Acc}
    & \textbf{F1} \\
\midrule
FinDialogLens (GPT-4o)
    & 92.1 & 94.3 & 94.5 \\
\makecell[l]{FinDialogLens (GPT-4o) +\\oracle message-level metadata}
    & 91.9 & 95.9 & 95.7 \\
\bottomrule
\end{tabular}
\caption{Ablation study on metadata error propagation.
We compare the original FinDialogLens pipeline using predicted
message-level metadata with an oracle setting that replaces predicted
metadata with gold message-level labels.}
\label{tab:metadata_error_propagation}
\end{table*}
The results suggest that error propagation from the message-level metadata module is limited. Replacing predicted metadata with gold metadata changes final price accuracy only marginally, from 92.1\% to 91.9\%, while improving trade outcome accuracy from 94.3\% to 95.9\% and trade outcome F1 from 94.5\% to 95.7\%. These results indicate that the downstream Trade Engine is relatively robust to the remaining errors produced by the high-performing Message-Level Module.

\section{Difficulty-Aware Router Analysis}
\label{appndix:cost}
\subsection{Routing Quality}
To further assess routing quality, we ranked all RFQ windows by their router logits and partitioned them into ten equal-sized bins. We then evaluated the accuracy of the rule-based engine within each bin to analyze the relationship between router confidence and routing effectiveness. As shown in Figure~\ref{fig:quality}, the top-ranked bins (0--10\%, 10--20\%, 20--30\%) achieved outstanding performance, with final price prediction accuracy ranging from 95\% to 100\%. In contrast, the lowest-ranked bin (90--100\%) exhibited a markedly lower accuracy of just 14.6\% in the RoBERTa-base router. These findings indicate that our fine-tuned routers are highly effective at identifying which RFQ events are most appropriate for rule-based processing.

\subsection{Cost and Latency}
We provide a detailed analysis of cost and latency for the routing experiments. The average cost per RFQ using GPT-4o is \$0.0054, with an average of 1,024.4 input tokens and 18.5 output tokens per RFQ. As shown in Table~\ref{tab:cost}, processing over 70,000 RFQs per day with the RoBERTa-base router results in daily savings exceeding \$300, while maintaining approximately 85\% accuracy for final price extraction.

For latency, the average response time for a single GPT-4o API call is 1.51 seconds per RFQ event. In our latency experiments, we measured the total processing time by sequentially submitting one RFQ event at a time, without batching or parallelization. The total latency for processing the entire test set with GPT-4o is approximately 1{,}459 seconds.

Table~\ref{tab:latency} reports total LLM API latency (in seconds) for the different router types under CPT(50\%) and CPT(80\%). Fine-tuned routers such as RoBERTa-base and Flan-T5-XL substantially reduce overall latency compared with the Random baseline. Under CPT(50\%), the RoBERTa-base router reduces total latency for final-price extraction to 217.4 seconds, compared with 717.8 seconds for the Random router. Flan-T5-XL achieves 234.9 seconds under CPT(50\%) for final price and 366.2 seconds for trade outcome, both well below the Random router's 760.1 seconds. These reductions support the practicality of our framework in deployment.

\begin{figure*}[t]
    \centering
    \includegraphics[width=1\textwidth]{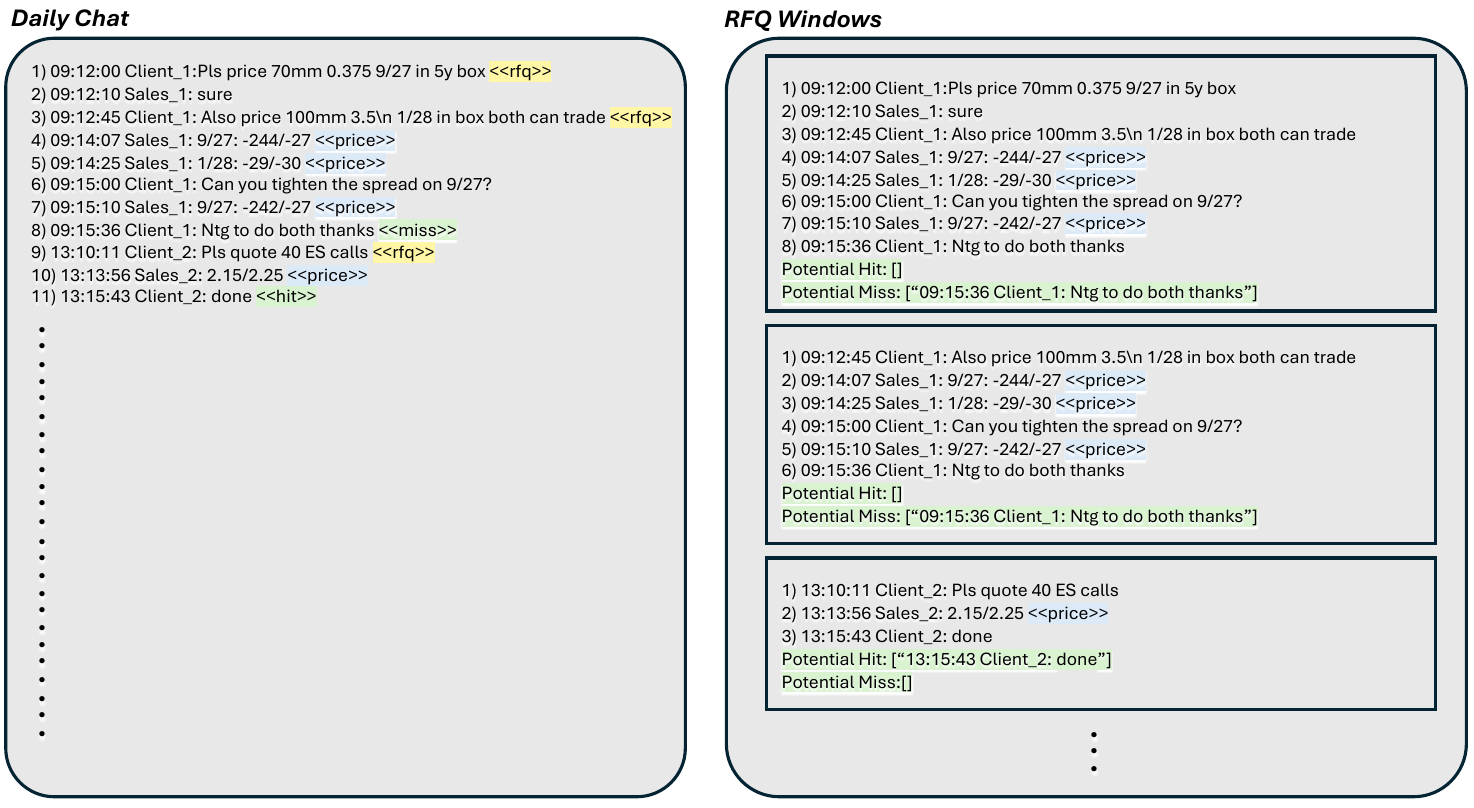}
    
\caption{Examples of the two types of \textbf{Input Context}: full daily chatroom and per-event RFQ window, supplied to the Trade Engine. Message-level metadata from the Message-Level Module is interleaved with the dialogue text using special tokens ($\ll$rfq$\gg$, $\ll$hit$\gg$, $\ll$miss$\gg$, $\ll$price$\gg$). CoT prompting baselines consume the full daily chatroom; FinDialogLens consumes per-event RFQ windows produced by the RFQ-Level Module.}
    \label{fig:input_context}
\end{figure*}

\begin{figure*}[t]
    \centering
    \includegraphics[width=0.9\textwidth]{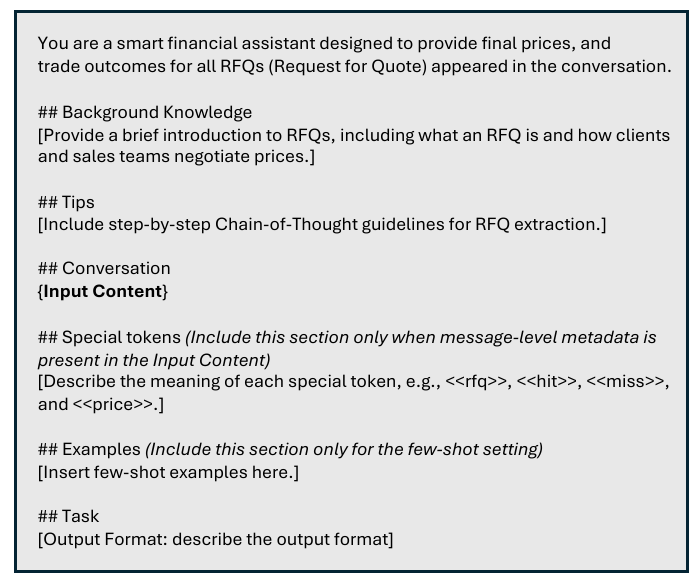}
    
    \caption{Prompt template for CoT prompting methods.}
    \label{fig:cot_prompt}
\end{figure*}

\end{document}